\documentclass[11pt,a4paper]{article}
\usepackage[hyperref]{acl2019}
\usepackage{times}
\usepackage{latexsym}
\usepackage{amsmath}
\usepackage{booktabs}
\usepackage{color}
\usepackage{graphicx}
\usepackage{latexsym}
\usepackage{svg}
\usepackage{textcomp}
\usepackage{times}
\usepackage{url}
\usepackage{tikz}
\usepackage{caption}
\usepackage{siunitx}
\newcommand{\cameraready}[1]{\textcolor{black}{#1}}

\aclfinalcopy 
\def\aclpaperid{***} 

\newcommand{\ELIF}{\textsc{ELI5}}
\newcommand*{\round}[1]{\num[round-mode=places,round-precision=1]{#1}}

\definecolor{light-blue}{rgb}{0.6,0.6,1}

\newcommand{\fair}{$^1$}
\newcommand{\loria}{$^2$}
\newcommand{\loriafair}{$^{1,2}$}
\newcommand{\nyu}{$^3$}
\newcommand{\google}{$^4$}

\title{ELI5: Long Form Question Answering}

\author{Angela Fan\loriafair{} \
  Yacine Jernite$^*$\fair{} \
  Ethan Perez$^*$\nyu{} \\
  \textbf{David Grangier\google{} \
  Jason Weston\fair{} \
  Michael Auli\fair{}} \\
  \fair{}Facebook AI Research \ \
  \loria{}LORIA \ \ \
  \nyu{}NYU $\ddagger$ \ \ \
  \google{}Google AI $\ddagger$  \\ 
  \tt{[angelafan,yjernite,jase,michaelauli]@fb.com,} \\
  \tt{perez@nyu.edu, grangier@google.com}
  }
\date{}

\begin{document}
\maketitle

\renewcommand* {\thefootnote}{\fnsymbol{footnote}}
\footnotetext{$^*$ Equal contribution}
\footnotetext{$\ddagger$ Work done while at Facebook AI Research}
\renewcommand*{\thefootnote}{\arabic{footnote}}

\begin{abstract}
We introduce the first large-scale corpus for long-form question answering, a task requiring elaborate and in-depth answers to open-ended questions.
The dataset comprises 270K threads from the Reddit forum ``Explain Like I'm Five'' (ELI5) where an online community provides answers to questions which are comprehensible by five year olds. 
Compared to existing datasets, ELI5 comprises diverse questions requiring  multi-sentence answers.
We provide a large set of web documents to help answer the question. 
Automatic and human evaluations show that an abstractive model trained with a multi-task objective outperforms conventional Seq2Seq, language modeling, as well as a strong extractive baseline. 
However, our best model is still far from human performance since raters prefer gold responses in over 86\% of cases, leaving ample opportunity for future improvement.\footnote{Dataset, Pretrained Models, and Additional Information is available: \url{https://facebookresearch.github.io/ELI5},  \url{https://github.com/facebookresearch/ELI5}}
\end{abstract}

\section{Introduction}

\begin{figure*}[th!]
    \centering
    \includegraphics[width=0.8\linewidth]{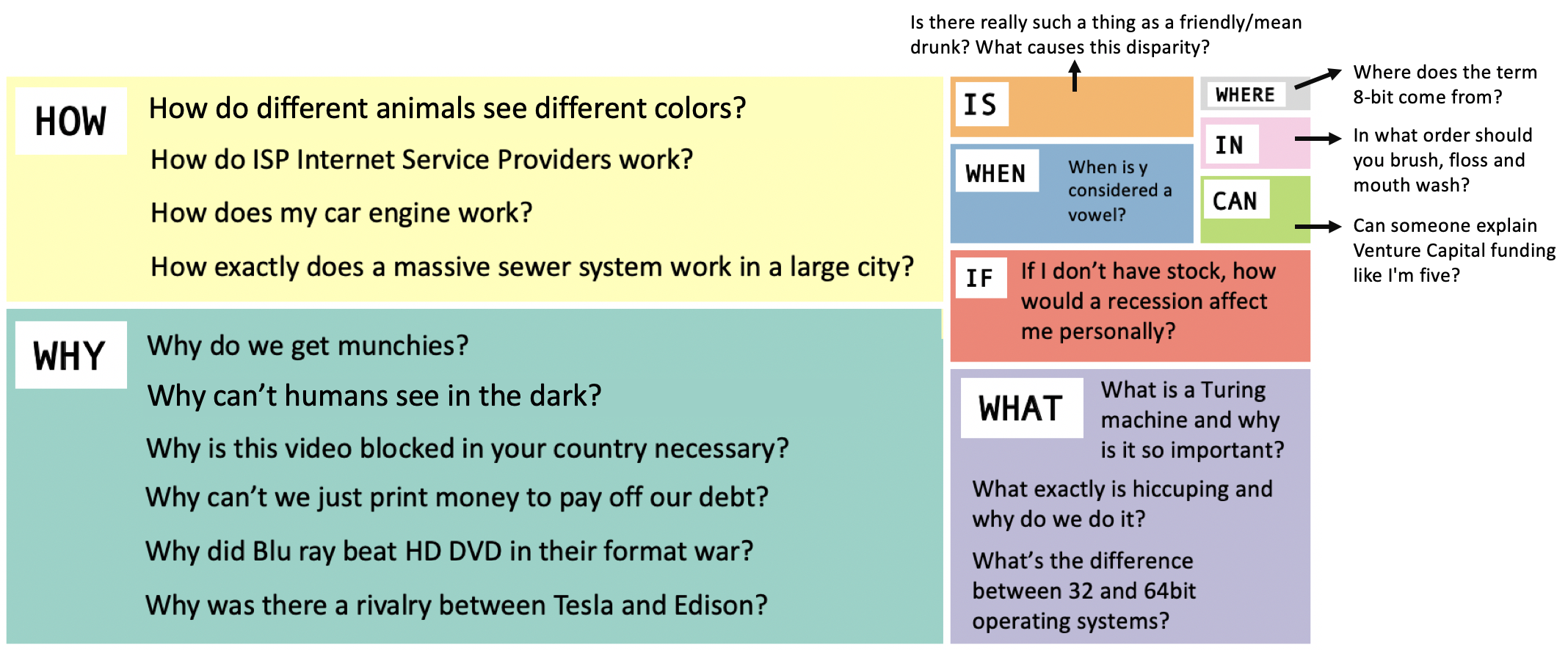}
    \caption{\ELIF~questions by starting word, where box size represents frequency. Questions are open ended and diverse.}
    \label{table:ELI5-questions}
\end{figure*}

Existing question answering datasets have enabled significant progress in models that provide extractive or unambigious short answers.
However, less attention has been paid to open-ended questions that require explanations.
In this work, we present ELI5: a Long Form Question Answering dataset that emphasizes the dual challenges of isolating relevant information within long source documents and generating paragraph-length explanations in response to complex, diverse questions (see illustrations in Figures~\ref{fig:example} and ~\ref{table:ELI5-questions}). 

\begin{figure}[ht]
  \small
  \rule{\linewidth}{1pt}
    \noindent \textbf{Question:} How do Jellyfish function without brains or nervous systems? [...] (60 words)\\[1.7mm]
    \noindent \textbf{Answer:} Jellyfish may not have a brain, but they have a rough nervous system and innate behaviours. However, they are very simple creatures. They're invertebrate: creatures without a backbone. Most jellyfish have really short life spans. Sometimes just a couple of hours. 
    [...] As their name implies, they are largely composed of basically jelly inside a thin membrane. They're over 95\% water. (327 words) \\[1.7mm]
    \noindent \textbf{Documents:} [...] Jellyfish do not have brains, and most barely have nervous systems. They have primitive nerve cells that help them orient themselves in the water and sense light and touch. [...] While they don’t possess brains, the animals still have neurons that send all sorts of signals throughout their body. [...] They may accomplish this through the assistance of their nerve rings. Jellyfish don't have brains, and that's just where things begin. They don't have many of the body parts that are typical in other animals. [...] (1070 words) \\[-1.5mm]
  \rule{\linewidth}{1pt}
  \caption{
    ELI5 example. Models must write multi-sentence answers given questions and supporting web documents.
    }
  \label{fig:example}
\end{figure}

The first challenge of ELI5 is the length and diversity of answers that span multiple sentences: questions are complex and cannot be easily addressed by a short response~\cite{nguyen2016ms} or by extracting a word or phrase from an evidence document \cite{rajpurkar2016squad}.
Answers also represent one of several valid ways of addressing the query. 
Many state-of-the-art question answering models perform well compared to human performance for extractive answer selection \cite{radford2018improving,devlin2018bert}.
However, their success does not directly carry over to our setting.

The second challenge is the length and diversity of the content from knowledge sources required to answer our questions. 
We leverage evidence queried from the web for each question.
In contrast to previous datasets where the human written answer could be found with lexical overlap methods \cite{weissenborn2017making}, ELI5 poses a significant challenge in siphoning out important information, as no single sentence or phrase contains the full answer. While there are some datasets that do require multi-sentence supporting knowledge such as  TriviaQA \cite{joshi2017triviaqa}, their answers are still short.

We benchmark the performance of several extractive, retrieval, and generative models. Evaluation of our task, and of multi-sentence text generation in general, is challenging.  We draw upon several evaluation metrics that quantify performance on intermediary fill-in tasks that lead up to the full answer generation. The overall answer generation quality is measured with ROUGE \cite{lin2004rouge} and various human evaluation studies. 

We develop a strong abstractive baseline by training a Seq2Seq model on multiple tasks over the same data: language modeling, masked word prediction \cite{devlin2018bert} and answer generation. 
We show this approach outperforms conventional Seq2Seq and language modeling, as well as a strong extractive baseline based on BidAF \cite{seo2017bidirectional} but generalized to multi-sentence output.
However, our best-performing model is still far from the quality of human written answers, with raters preferring the gold answers 86\% of the time.
Further, we show that model performance is strongly limited by the ability to comprehend long multi-document input and generate long outputs to form a comprehensive answer, leaving this challenge for future research.
\section{Related Work}
\label{sec:Related Work}

\begin{table*}[t]
  \centering
  \resizebox{\textwidth}{!}{%
    \begin{tabular} {c|ccc|cccccccc|c}  \toprule
      Dataset                                          & \multicolumn{3}{c|}{Average \# of Words}        & \multicolumn{8}{c|}{1st Question Word Frequency (\%)}                                                                     & \multicolumn{1}{c}{} \\
                                                       & Question       & Document(s)   & Answer         & Why           & How           & What          & When          & Where        & Who           & Which        & OTHER       & \# Q-A Pairs  \\
      \midrule 
      \textbf{\ELIF}                                   & \textbf{42.2}  & 857.6 (212K)  & \textbf{130.6} & \textbf{44.8} & \textbf{27.1} & 18.3          & \textbf{11.3} & 2.0          & 1.8           & 0.8          & 6.1         & \textbf{272K} \\
      \midrule 
      MS MARCO v2 \break \small{\citep{nguyen2016ms}}  & 6.4            & 56            & 13.8           & 1.7           & 16.8          & 35.0          & 2.7           & 3.5          & 3.3           & 1.8          & 35.3        & 183K          \\
      TriviaQA \break \small{\citep{joshi2017triviaqa}}& 14             & \textbf{2895} & 2.0            & 0.2           & 3.9           & 32.6          & 2.0           & 2.1          & 16.8          & \textbf{41.8}& 0.6         & 110K          \\
      NarrativeQA \break \small{\citep{kocisky2018the}}& 9.8            & 656           & 4.7            & 9.8           & 10.7          & 38.0          & 1.7           & \textbf{7.5} & \textbf{23.4} & 2.2          & 6.8         & 47K           \\
      CoQA \break \small{\citep{reddy2018coqa}}        & 5.5            & 271           & 2.7            & 2             & 5             & 27            & 2             & 5            & 15            & 1            & \textbf{43} & 127K          \\
      SQuAD (2.0) \break \small{\citep{rajpurkar2018know}}  & 9.9            & 116.6         & 3.2            & 1.4           & 8.9           & \textbf{45.3} & 6.0           & 3.6          & 9.6           & 4.4          & 17.6        & 150K\\
      HotpotQA  \break \small{\citep{yang2018hotpotqa}}& 17.8           & 917           & 2.2            & 0.1           & 2.6           & 37.2          & 2.8           & 2.2          & 13.8          & 28.5         & 12.8        & 113K\\
      \bottomrule
    \end{tabular}
  }
  \caption{
    Comparing large-scale QA datasets. \ELIF~has answers an order of magnitude longer and more open-ended questions.
  }
  \label{tab:ELI5:qa}
\end{table*}

Various QA datasets have been proposed in roughly two categories: extractive answers and short abstractive answers (see Table~\ref{tab:ELI5:qa}).

\paragraph{Extractive QA} Extractive question answering datasets such as 
TREC~\cite{voorhees2003overview},
SQuAD~\cite{rajpurkar2016squad,rajpurkar2018know}, NewsQA~\cite{trischler2017newsqa}, SearchQA~\cite{dunn2017searchqa}, and 
QuAC~\cite{choi2018quac} constrain the answer to a word or short phrase from the input  and evaluate using exact match or F1 with the ground truth span. 
HotpotQA~\cite{yang2018hotpotqa} extends this approach by building questions which  challenge models to conduct multi-hop reasoning across multiple paragraphs, but the answer is still a short span. Further, the answer must be 
straightforward,
as it needs to be copied from the supporting evidence --- precluding most ``how'' or ``why'' type questions.

\paragraph{Abstractive QA}  Abstractive datasets include NarrativeQA~\cite{kocisky2018the}, a dataset of movie and book summaries and CoQA~\cite{reddy2018coqa}, a multi-domain dialogue dataset. Both collect responses with crowdworkers and find that written answers are mostly extractive and short. MS MARCO~\cite{nguyen2016ms}, a dataset of crowdsourced responses to Bing queries, has written answers around 1 sentence long with short input passages. 
TriviaQA~\cite{joshi2017triviaqa} contains longer multi-document web input, collected using Bing and Wikipedia. As the dataset is built from trivia, most questions can be answered with a short extractive span. 

\paragraph{Multi-document summarization} The \textsc{ELI5} task of writing a paragraph length response from multiple supporting documents can be seen as a form of query-based multi-document summarization~\cite{tombros1998advantages}. Summarization tasks such as DUC 2004\footnote{\url{https://duc.nist.gov/duc2004/}} involve long input and multi-sentence generation, but contain much less training data compared to ELI5. 
WikiSum \cite{liu2018generating} proposes writing Wikipedia articles as a multi-document summarization task. 
ELI5 requires more directed text generation to answer a question, rather than to write about a general topic. In addition, ELI5 contains a diverse set of questions which can involve more than one Wikipedia concept.

\section{Making a Long Form QA Dataset}
\label{sec:ELI5}

\subsection{Creating the Dataset from ELI5}
\label{sec:ELI5:creation}

There are several websites which provide forums to ask open-ended questions such as Yahoo Answers, Quora, as well as numerous Reddit forums, or subreddits. 
We focus on the subreddit \textit{Explain Like I'm Five} (ELI5) where users are encouraged to provide answers which are comprehensible by a five year old.\footnote{\url{https://www.reddit.com/r/explainlikeimfive}}
ELI5 is appealing because answers \cameraready{are supposed to be entirely self contained, and thus} rely less on pre-existing knowledge of the world and use simpler language that is easier to model.

\paragraph{Questions and answers.} 
We select a set of questions and answers from the ELI5 forum up to July 2018 and then filter it based on how users rated these pairs.
First, we only retain questions which have a score of at least two, that is two more `up-votes' than `down-votes'.
Second, there must be at least one answer with a score of at least two.
This yields a final number of 272K questions, and ensures that at least one person other than the author has read the thread and deemed it appropriate.
For each thread, we select the answer with the highest voting score as the reference. 
Note that 63\% have one or more other valid answers \cameraready{by our up-vote criteria}, potentially doubling the size of the available training data.

\paragraph{Preparing supporting information.} 
Next, we collect web sources for every question to provide relevant information that a system can draw upon when generating an answer. 
Wikipedia has been found effective for factoid-oriented questions \citep{joshi2017triviaqa, chen2017reading}. However, early experiments in our setting showed it to be insufficient to cover the wide range of topics present in ELI5 and to address the open-ended nature of the questions. 
Instead, we use web data provided by Common Crawl.\footnote{\url{http://commoncrawl.org}}
Specifically, we consider each of the individual pages in the July 2018 archive 
(roughly one per URL) 
as a single document. 
The data is tokenized with Spacy\footnote{\url{https://spacy.io}} and we select English documents with FastText language identification \citep{bojanowski2017}. 
Finally, we index the data with Apache Lucene.\footnote{\url{http://lucene.apache.org}}

\paragraph{Creating support documents.} 
We query the index for the 272K questions and gather the 100 most relevant \textit{web sources} for each question, excluding Reddit. Each \textit{web source} is the extracted text of one page in Common Crawl.  
This leads to supporting text for each question of a few hundred thousand words. 
There is a good chance that the supporting text contains the necessary information to answer the question, but the sheer amount of data is far beyond the scope of what many modern models can handle. 
We therefore filter the 100 web sources by selecting specific passages using a simple heuristic: we split each web source into sentences, find sentences with the highest TFIDF similarity with respect to the question, add some local context for each of these, and concatenate the result into a single \textit{support document}, with special tokens indicating non-contiguous passages and document shifts. Each \textit{support document} is the result of this processing to concatenate relevant information from the web sources. 

We find that extracting 15 passages with a context of one sentence before and after the initial selection provides the best trade-off between support document length and likelihood of containing relevant information, where relevance is measured as the likelihood of containing a sentence which has high ROUGE with the answer.
We release all 100 Common Crawl IDs for each question and a script to create the support document 
so future research can use the support document or choose to further investigate the information retrieval problem. 

\paragraph{Finalizing the data set.} 
If the training data contains questions that are too similar to the validation and test data, a model may perform well on these examples by memorizing related examples.
We prevent this by building the validation and test set to contain questions that are sufficiently different from the training data. 
We compute the TFIDF similarity between each pair of questions in the entire dataset and sample the validation and test set from the subset which has no close neighbor by TFIDF score. The final dataset contains 237K train examples, 10K for valid, and 25K for test.

\subsection{Dataset Analysis}

\begin{table}[t]
   \centering \footnotesize
\begin{tabular}{l|c}
    \toprule 
     \% Correct Human Answers & 94.5 \\
     \% Correct Human Answers with Explanation & 90.2 \\ 
     \\
     \% Support Document contains Full Answer & 65.0 \\
     \% Support Document contains Relevant Info & 92.0 \\ 
    \bottomrule 
 \end{tabular}
    \caption{Annotated subset of ELI5 to assess answerability.}
  \label{tbl:human_annotation_data_quality}
 \end{table}

Table~\ref{tab:ELI5:qa} compares ELI5 to related datasets in terms of the length of the question, support document, answer, as well as statistics on the question types. 

First, ELI5 questions are much longer than in other datasets. This is because the initial question is often followed by a clarifying paragraph detailing what aspect of the general theme should be addressed or the question's starting assumptions, which need to be considered to answer well. 
To get a rough idea of the different questions, we categorize them based on interrogative words. 
ELI5 focuses on open-ended queries which are less represented in other extractive or abstractive datasets.
Figure \ref{table:ELI5-questions} shows examples of ELI5 questions split by type and Appendix Figure~\ref{fig:example_qa_pairs} displays random examples from the ELI5 training set. 
Interestingly, even \textit{What} questions tend to require paragraph-length explanations (\textit{What is the difference\ldots}).

Support documents contain $22$-$60$ sentences or on average $858$ words, which puts ELI5 on the higher end of published datasets for document length.
ELI5 contains long-form answers with an average length of $6.6$ sentences, or $130$ words.

Next, we analyze a random subset of ELI5 to assess the feasability of answering the questions in the dataset.
We judge \cameraready{if the question is answerable} by reading each question, the gold answer, and the support document we have created with TF-IDF extraction. 
\cameraready{Note that} questions can have multiple parts and all parts of the question must be answered.
\cameraready{We sample} 500 randomly question-answer pairs from the training set \cameraready{and find that} 94.5\% of gold answers fully address the question (Table~\ref{tbl:human_annotation_data_quality}) \cameraready{based on the information in the support document.}
Figure~\ref{fig:example_non_answered} in Appendix~\ref{appendix:Examples} displays examples of human answers that do not correctly answer the question.  
A small proportion of answers are correct but do not explain the answer. On the support \cameraready{document} side,
65\% of the \cameraready{support} documents we construct provide the complete answer to the question, and 92\% of support documents provide information relevant to the question. 
 
\begin{figure*}[t!]
  \footnotesize
  \rule{\linewidth}{1pt}
    \noindent \textbf{Extractive:} But none of these hypotheses make much sense for a jellyfish because they do n’t have brains at all. They just have a nerve net --- a loose ring of neurons that runs around the rim of their pulsating bells. They have an unusual nervous system, because jellyfish are not bilaterally symmetrical --- that is, they don’t have a left side and a right side. Jellyfish don't have brains, but their nervous systems detect smells, light and other stimuli, and they coordinate their physical responses.\\[1.7mm]
    \noindent \textbf{Language Model:} They don't have brains. They have a nervous system. They do not have a brain. They are not conscious. They can not move. They move. Their nervous system is not a muscle. They use their muscles to move. The nervous system does not have any nerves. They just have a nerve. They also have a muscle that is very similar to a muscle in that it is very much like a muscle, but instead of a muscle it is a muscle and it is like a nerve, but it is not muscle. The nerve is like the muscle, it is just a muscle but it has a nerve and it has nerves. The nerves are like the muscles, but they have nerves. \\[1.7mm]
    \noindent \textbf{Multi-task Seq2Seq:} Jellyfish don't have brains. Jellyfish have a nervous system. They know how to swim, they don't even know what to do with their bodies. They have a ``nervous system'', which is basically a bunch of nerve cells that connect to the body. The nervous system is made up of neurons, neurons, nerves, etc. Each neuron doesn't have a specific set of ``neurons'', it just has a different set of nerves. It has a set of sensory neurons, which are all connected to each other. It just knows that it does react to the stimulus, and it knows that the stimulus is there. They have a bunch of nervous systems that act like a filter to get information back.\\[-1.5mm]
  \rule{\linewidth}{1pt}
  \caption{Example answers from the extractive BidAF model, Question + Document + Answer language model, and Multi-task Seq2Seq model for the question ``How do Jellyfish function without brains or nervous systems?'' (cf. Figure~\ref{fig:example}).
    }
  \label{fig:example_answers}
\end{figure*}

\section{Evaluation Methods}

\paragraph{Evaluating long-form answers.} There are several aspects to quality: answers should be topical and accurate, fluent, and coherent from start to end. We judge the accuracy aspect by comparing to the gold answer.
\textsc{ROUGE} \cite{lin2004rouge} measures similarity between a model output and one or several references, and is often used in summarization. 
While our task presents different challenges, such as the diversity of possible answers to a question, we still find the corpus-level metric to be useful to rank different related models (\textsection\ref{sec:results}). 
We report F1 for \textsc{ROUGE-1}, \textsc{ROUGE-2}, and \textsc{ROUGE-L}. 

\paragraph{Abstractive model metrics.} 
For generative models, perplexity (\textsc{PPL}) measures the ability to predict the next word in a sequence given its context.
For a variant which focuses on semantically important words, we  report \textsc{FILL-1}, the accuracy at which models generate different Nouns, Verbs, and Adjectives given the correct preceding tokens in the first 2K examples of the test set.
Finally, \textsc{ROUGE-20\%} measures the model's ability to complete an answer given the first 80\% of the reference answer, the question, and the support document. Specifically, we generate a number of tokens corresponding to 20\% of the average answer length in the validation set, and measure \textsc{ROUGE} between these and the last 20\% of the reference. 
We mentioned that there are several valid ways to answer most questions.
This measure abstracts away this variability and evaluates a system's ability to complete an answer.

\paragraph{Human evaluation.} 
We use crowdworkers to conduct three assessments. First, evaluators rate the \textit{fluency} of human and model generated answers on a 5-point Likert Scale, from ``very poorly written" to ``easily readable'' (500 evaluations). 
Second, evaluators are given question-answer pairs and are asked if the answer is \textit{correct} (500 evaluations) \footnote{\cameraready{We experimented with a variant where crowdworkers were allowed to select a third \textit{I don't know} option, but found it was used only around 8\% of the time.}}. We also evaluated a smaller subset ourselves while additionally looking at the support documents (100 evaluations) to assess answer accuracy.
Lastly, crowdworkers are given the question and answers from two models and asked to decide which answer they \textit{prefer} while considering readability and accuracy (1000 evaluations). 
Each crowdworker assessment is made by 3 different evaluators. The same questions are used for all models and must be at least 5 words long.

\section{Models}

\subsection{Extractive and Retrieval Models}

\paragraph{Retrieval baseline and oracle.} We report \textsc{ROUGE} for a retrieval system that returns the answer of the closest question in the training set. 
Specifically, we perform a nearest neighbor search~\citep{johnson2017} over the average word embeddings of the question using \textsc{fasttext}~\cite{bojanowski2017}. 
We also compute an approximate oracle score for extractive systems by using the reference answer to select similar sentences from the support document to maximize \textsc{ROUGE}.
Computing \textsc{ROUGE} between the reference and all sets of sentences from the source is intractable. 
Instead, we perform a beam search that adds sentences maximizing TFIDF with respect to the answer. 
The final beam is re-ranked using \textsc{ROUGE} with respect to the reference answer. 
We run this algorithm on our support document and on the full set of web sources for each validation and test question, selecting up to 10 sentences with a beam of size 10. 

\paragraph{Extractive models.} The first baseline we explore simply returns the $7$ sentences from the support document which have the highest TFIDF similarity with the question. We also evaluate models which score sentences from the support document based on the question and return the highest scoring sentences in their original order (the number is tuned on the validation set to maximize \textsc{ROUGE}). We  train a model based on BidAF  \cite{seo2017bidirectional}. We create an extractive training set by finding the span of up to $5$ contiguous sentences in the support document which have the highest \textsc{ROUGE} with respect to the reference answer, and sub-sample other support document sentences so that the final training document is shorter than $400$ words. We then train a BidAF model to predict the extracted span in the sub-sampled support document based on the question. For test, we compute the span score for each individual sentence, and return the $5$ with the highest score as it performed best compared to returning 3 or 7 sentences.

\subsection{Abstractive Models}
\label{sec:abstractive}

\paragraph{Language and Seq2Seq models.} We train several models based on the Transformer architecture \cite{vaswani2018attention}, both in its language model and sequence-to-sequence (Seq2Seq) configurations.
To investigate how much information from the document the model uses, we train a language model on the concatenation of \textit{Question}, \textit{Support Document}, and \textit{Answer} (Q + D + A) as well as on the \textit{Question} and \textit{Answer} (Q + A). Similarly, one Seq2Seq configuration goes from Q to A, and the other from Q + D to A. In all cases, Q, D, and A are separated by special tokens.

\paragraph{Multi-task training.} 
Language models are trained to predict all tokens in the question, web source, and answer. However, the standard Seq2Seq model only receives training signal from predicting the answer which is much less than the language model gets.
This can contribute to learning poor quality representations compared to language models. 
To address this, we train a \textit{multi-task} Seq2Seq model: during training, we multi-task between several generation tasks, including language modeling of Q + D + A by the decoder and variations of source/target pairs (see Appendix~\ref{appendix:multitask}).
We add a masked word prediction task \cite{devlin2018bert} where 15\% of tokens in the input are masked
and must be recovered by the model in the correct order,
and append a marker at the start of each sequence to indicate the task.

\begin{table}[t!]
  \centering \small
  \begin{tabular}{ l c c  c  c }\toprule
    \bf{Model} & \textbf{PPL} & \multicolumn{3}{c}{\bf{ROUGE}}\\
               &              & \bf{1} & \bf{2} & \bf{L}\\\hline\hline
        Support Document            &   -   & \round{16.83} & \round{2.32} & \round{10.17} \\
        Nearest Neighbor            &   -   & \round{16.74} & \round{2.31} & \round{12.45} \\ 
        \midrule 
        Extractive (TFIDF)          &   -   & \round{20.57} & \round{2.85} & \round{17.0}  \\
        Extractive (BidAF)          &   -   & \round{23.50} & \round{3.11} & \round{17.45} \\ 
        Oracle support doc            &   -   & \round{27.40} & \round{2.82} & \round{19.87} \\
        Oracle web sources          &   -   & \round{54.77} & \round{8.61} & \round{40.33} \\
        \midrule
        LM Q + A                    & \round{42.21} & \round{27.77} & \round{4.67} & \round{23.12} \\
        LM Q + D + A                & \round{33.85} & \round{26.44} & \round{3.96} & \round{20.52} \\
        Seq2Seq Q to A              & \round{52.87} & \round{28.28} & \round{5.07} & \round{22.72} \\
        Seq2Seq Q + D to A          & \round{55.05} & \round{28.32} & \round{5.11} & \round{22.75} \\
        Seq2Seq Multi-task          & \round{32.69} & \round{28.94} & \round{5.44} & \round{23.12} \\ \bottomrule
\end{tabular}
   \caption{Comparison of oracles, baselines, retrieval, extractive, and abstractive models on the full proposed answers.}
 \label{tbl:full_rouge_answer}
\end{table}

 \begin{table}[t]
   \centering \small
   \begin{tabular}{ l c @{\hspace{1\tabcolsep}} c @{\hspace{1\tabcolsep}} c c @{\hspace{1\tabcolsep}} c @{\hspace{1\tabcolsep}} c c }\toprule
     \bf{Model}              & \multicolumn{3}{c}{\bf{FILL-1 acc.}}  & \multicolumn{3}{c}{\bf{ROUGE-20\%}} \\
                             & \textbf{N} & \textbf{V} & \textbf{A}  & \bf{1} & \bf{2} & \bf{L}  \\\hline\hline
         LM Q + A            & 31.0       & 29.6       & 20.6        & 26.5  & 7.0   & 21.1 \\
         LM Q + D + A        & 30.9       & 28.9       & 19.9        & 26.3  & 7.8   & 21.3 \\
         S2S Q to A          & 21.7       & 23.0       & 15.5        & 33.6  & 11.5  & 29.5 \\
         S2S Q + D to A      & 27.6       & 26.3       & 19.4        & 32.7  & 10.7  & 28.6 \\
         S2S Multi-task      & 27.9       & 26.7       & 19.9        & 37.2  & 14.6  & 33.0 \\ \bottomrule
 \end{tabular}
    \caption{Intermediary fill-in tasks for sequential generation.}
  \label{tbl:fill_rouge}
 \end{table}
 
\paragraph{Data processing.} To reduce the vocabulary, we apply byte-pair encoding \cite{sennrich2016neural} to generate 40K codes which are applied to all datasets.
We model a vocabulary of 52,863 tokens for answer generation. We use the Transformer implementation of fairseq-py~\cite{gehring2017convolutional} and train with the big architecture following the details in \cite{vaswani2018attention}. Given our data length, we train with a large batch size by delaying gradient updates until a sufficient number of examples have been seen \cite{ott2018scaling}. 

\paragraph{Generation.} We generate from abstractive models using beam search with beam 5. We disallow repeated trigrams to prevent repetition, a technique commonly used in multi-sentence summarization \cite{paulus17, fan2018controllable}. For the full answer generation task, we tune a minimum and maximum length for generation on the valid set and apply these settings to the test set.

 \section{Results}
\label{sec:results}

\begin{figure*}
    \centering
    \includegraphics[width=0.8\linewidth]{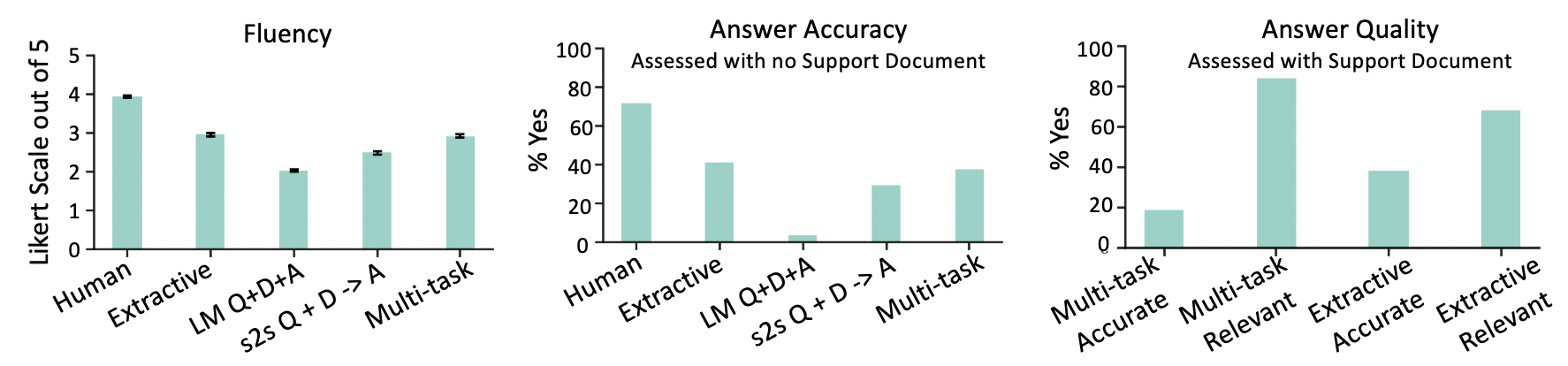}

    \caption{Human evaluation of answer fluency and accuracy --- with and without access to supporting evidence documents}
    \label{fig:human_fluency2}
\end{figure*}

\begin{figure}[t!]
    \centering
    \includegraphics[width=.7\linewidth]{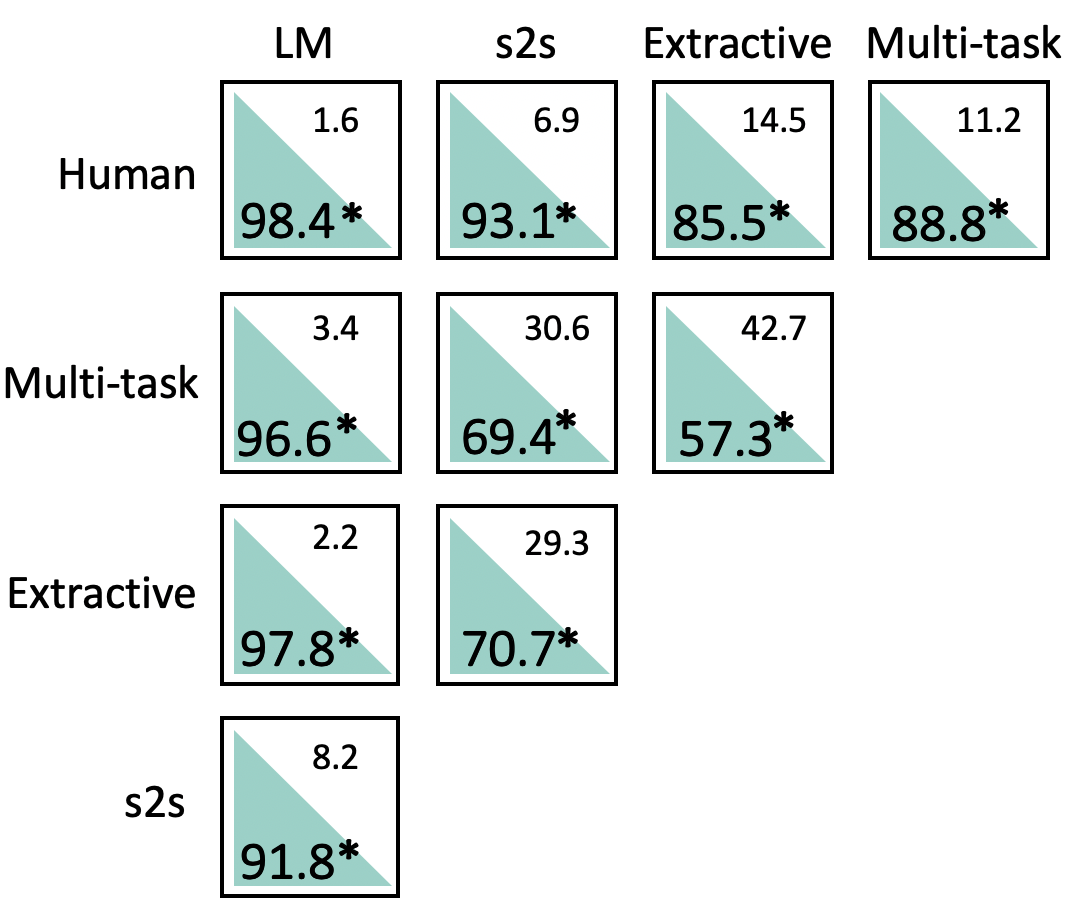}
    \caption{Human preferences for pairwise comparisons. The better model's \% preference is bolded. * indicates statistical significance.}
    \label{fig:human_eval_comparison}
\end{figure}

\subsection{Overview of Model Performance}

\begin{figure*}[t]
    \centering
    \includegraphics[width=\linewidth]{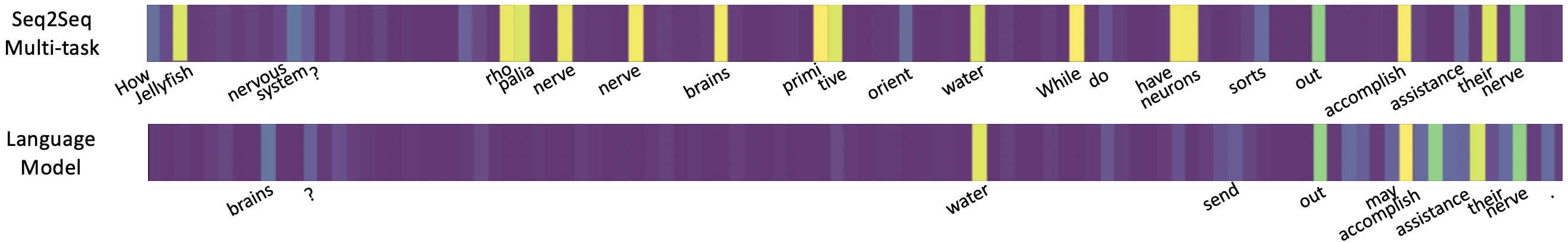}
    \caption{Attention over the question and supporting evidence for the Multi-task Seq2Seq model and Question + Document + Answer language model. Attention is shown for the first word of answer generation.}
    \label{fig:attn_comparison}
\end{figure*}

\paragraph{Full answer \textsc{ROUGE}.} 
Table~\ref{tbl:full_rouge_answer} shows that the nearest neighbor baseline performs similarly to simply returning the support document which indicates that memorizing answers from the training set is insufficient. 
For extractive models, the oracle provides an approximate upper bound of $27.4$ \textsc{ROUGE-1}. The BidAF model is the strongest ($23.5$), better than TFIDF between the question and the support document to select sentences. However, these approaches are limited by the support document, as an oracle computed on the full web sources achieves $\round{54.77}$.

Abstractive methods achieve higher \textsc{ROUGE}, likely because they can adapt to the domain shift between the web sources and the ELI5 subreddit. In general, Seq2Seq models perform better than language models and the various Seq2Seq settings do not show large \textsc{ROUGE} differences. Figure~\ref{fig:example_answers} shows an example of generation for the language model and the best Seq2Seq and extractive settings (see Appendix~\ref{appendix:Examples} for additional random examples).

\paragraph{Perplexity and fill-in tasks.} Tables~\ref{tbl:full_rouge_answer} and~\ref{tbl:fill_rouge} present metrics specific to sequential generation models: perplexity of the answer, accuracy of the model's \textsc{FILL-1} word prediction for Nouns, Verbs, and Adjectives, and \textsc{ROUGE} of the conditional generation of the last $20\%$ answer words. The language model perplexity is much lower than that of the standard Seq2Seq setting -- this is likely linked to the number of output tokens the system is required to predict at training time. The multi-task Seq2Seq experiment, in which the Seq2Seq decoder is trained to predict the question and the document, in addition to the answer, can reach the same perplexity as the language model.
\textsc{ROUGE-20\%} shows a much starker contrast between language modeling and Seq2Seq, as well as between standard Seq2Seq and multi-task training. The latter achieves strong performance of $\round{37.18}$ \textsc{ROUGE-1}. However, both versions of the language model are still better at FILL-1. These results suggest that the Seq2Seq model is better than the language model in maintaining coherence and that Seq2Seq relies on information over many time steps.

\paragraph{Human evaluation.} 
Human answers are rated highest in terms of fluency (Figure~\ref{fig:human_fluency2}, left). 
The extractive model outputs human-written text which is likely fluent but with the  failure mode of concatenating unrelated sentences. 
The multi-task model performs similarly to the extractive model which indicates that abstractive methods can generate coherent answers. The language model and standard Seq2Seq trail behind.

To get a sense of the stability of our results, we analyzed the standard deviation of three independent fluency trials conducted on separate days and we find low variation (Appendix~\ref{sec:humanvariance}, Figure~\ref{fig:human_variation_appendix}). 
We also measure agreement between crowdworkers in selecting positive (scores 4 and 5), negative (1 and 2), or neutral (3) choices on the 5-point Likert scale, and find that 2 crowdworkers agree almost 100\% of the time (Appendix~\ref{sec:humanvariance},  Figure~\ref{fig:human_variation_appendix}). 

In answer accuracy (Figure~\ref{fig:human_fluency2}, middle), there is a large gap between human performance and all models. The language model is almost never accurate, while the extractive model is slightly more so than the multi-task model. 
Crowdworkers assessing accuracy do not have the support document. We evaluate accuracy ourselves with the support document in Figure~\ref{fig:human_fluency2}, right. Similar to crowdworkers, we find 40\% of extractive answers to be accurate. We find only 19\% of multi-task model answers are fully accurate; even if the model output answers the question, it can generate a sentence with an incorrect statement. In contrast, the extractive model copies sentences from human-written text. However, the multi-task model is better at generating relevant answers (84\% relevancy compared to 68\% for extractive), as the extractive model is constrained by the support document.

Figure~\ref{fig:human_eval_comparison} presents pairwise preference judgments of human annotators shown answers from two of the five systems. The reference answer is preferred over the output of all of our trained models in at least $85.5\%$ of cases, indicating there is substantial room for improvement. The multi-task abstractive setting comes next, closely followed by the extractive (multi-task is only preferred in $57\%$ of comparisons), then the standard Seq2Seq and finally the language model, considered worse than any other setting in at least $91\%$ of cases. 

We use a two-tailed binomial test to test statistical significance of the pairwise judgments and it shows that all judgments are statistically significant at $p < 0.05$. 

\subsection{Quantitative and Qualitative Analysis}

\paragraph{Discussion of the proposed metrics.} 
We present a number of metrics which provide insight into various model behaviors. We recommend future work to report full \textsc{ROUGE} and \textsc{ROUGE-20\%}.
Perplexity and \textsc{FILL-1} focus on local prediction and are poor indicators of overall appropriateness for the full task. Full answer \textsc{ROUGE} discriminates reasonably well between models with the same general architecture, but cannot rate an abstractive system against an extractive one. The \textsc{ROUGE-20\%} measure abstracts away some variability and focuses on coherence between the beginning and end of an answer. This metric correlates with human judgments of quality but can only be reported for sequential generation. 

\paragraph{Analysis of extractive, LM and Seq2Seq models.} 
Language models perform better than Seq2Seq in terms of perplexity and \textsc{FILL-1}, while being significantly worse at \textsc{ROUGE-20\%} and human evaluations. To investigate this, we visualize the attention mechanism at the start of answer generation in Figure~\ref{fig:attn_comparison}. The attention of the language model is strongly focused on nearby context when generating the first word of the answer, whereas the multi-task Seq2Seq model attends more evenly to relevant information in the question and the document. This validates our assumption that the language model's focus on local context is insufficient for high quality answers. 

\begin{figure}[t!]
    \centering
    \includegraphics[width=\linewidth]{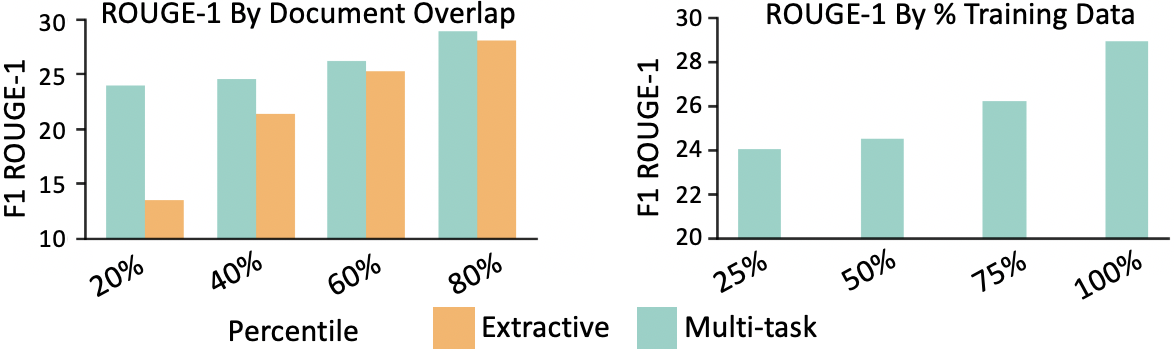}
    \caption{(left) Model score by document-answer similarity. (right) Seq2Seq multi-task score by amount of training data.}
    \label{fig:doc_overlap}
\end{figure}

In Figure~\ref{fig:doc_overlap} (left), we further investigate how the relevance and quality of the support document extraction step affects the answers provided by the extractive and abstractive setting. The \textsc{ROUGE} score is displayed for data subsets, partitioned by percentile of word overlap of the answer with the support document (e.g. how many answer words appear). While both models perform better for documents with higher \textsc{ROUGE} overlap between support document and human answer, the abstractive setting is much better at compensating for when the support document has lower relevance.

\begin{figure}[t!]
    \centering
    \includegraphics[width=1\linewidth]{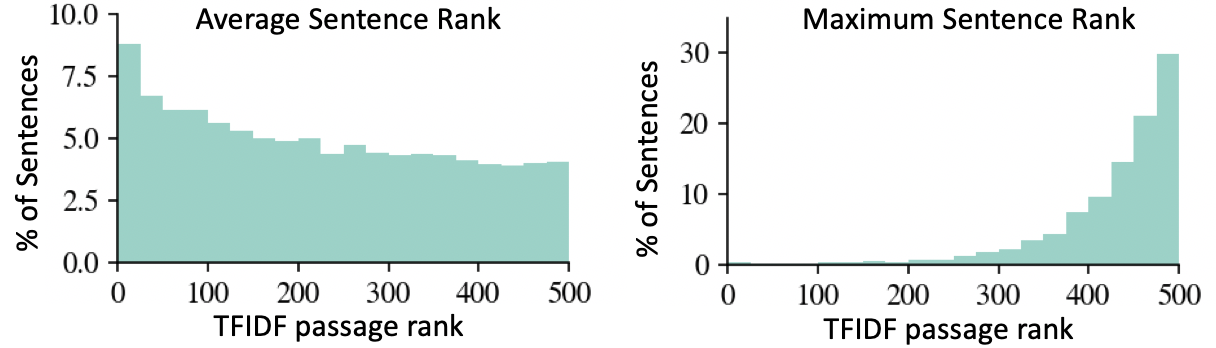}
    \setlength{\abovecaptionskip}{0.6pt}
    \caption{(left) TFIDF rank of source passage for oracle sentences. (right) Highest rank used per question.}
    \label{fig:tfidf_oracle}
\end{figure}

\paragraph{Data size and initial selection.} 
There is a large difference between the extractive oracle \textsc{ROUGE} using our support document and the oracle on full web sources.
This suggests that the initial selection of our support document severely limits access to relevant information.
To assess the impact of support document size,
we re-run the selection step for $1000$ examples to extract $500$ passages instead of $20$,
and run the oracle on these new inputs. Figure~\ref{fig:tfidf_oracle} shows the TFIDF rank of the passages from which sentences are selected. 
While slightly more sentences are extracted from the higher ranking passages, less than $9\%$ come from the first $20$, and most oracles have at least one sentence from the last $100$. For a model to perform best, it would have to handle inputs tens of thousands of words long. In Table~\ref{tbl:full_rouge_answer}, we show an oracle computed on the full web sources has much higher ROUGE than an oracle computed on the support document.

We analyze the impact of data size on performance in Figure~\ref{fig:doc_overlap}. We train the multi-task model on 25\%, 50\%, and 75\%, and the all of the data to compare performance.
\textsc{ROUGE} increases as a function of the data used and even though
ELI5 is one of the larger QA datasets (\textsection\ref{sec:ELI5}), this shows that collecting more still helps. While we only used one reference answer per question here, recall that over half of them have multiple answers, which could be leveraged to train better models.

\cameraready{\paragraph{Combining challenges.} 
Our task blends the inter-dependent challenges of retrieving information, reasoning, and writing long outputs. Studying each of these aspects in context is particularly important. For example, we show that the abstractive model's ability to compensate for a (realistically) imperfect support document is essential to its relative success over extractive methods. The fluency gap between the reference and the extractive system in human evaluation also suggests that the latter may require sequential decision capabilities. This kind of decision making is necessary to address the dual challenges of reasoning over several supporting facts and generating long coherent outputs. We see our task's need to combine complementary systems as critical to gaining insights into their individual behaviors.}

\section{Conclusion}
\label{sec:Conclusion}

We introduce the first large-scale long form question answering dataset of open-ended queries with explanatory multi-sentence answers.
We show that abstractive models generate coherent answers and are competitive with extractive models in human evaluation. 
Proposed models are far from human performance, in part due to the inability to exploit the 
long full web text.
We hope ELI5 will inspire future work in all aspects of long-form QA, \cameraready{from the information extraction problem of obtaining information from long, multi-document input to generating more coherent and accurate paragraph-length answers.}

\bibliography{master}

\begin{thebibliography}{27}
\expandafter\ifx\csname natexlab\endcsname\relax\def\natexlab#1{#1}\fi

\bibitem[{Bojanowski et~al.(2017)Bojanowski, Grave, Joulin, and
  Mikolov}]{bojanowski2017}
Piotr Bojanowski, Edouard Grave, Armand Joulin, and Tomas Mikolov. 2017.
\newblock Enriching word vectors with subword information.
\newblock \emph{{TACL}}, 5:135--146.

\bibitem[{Chen et~al.(2017)Chen, Fisch, Weston, and Bordes}]{chen2017reading}
Danqi Chen, Adam Fisch, Jason Weston, and Antoine Bordes. 2017.
\newblock \href {https://doi.org/10.18653/v1/P17-1171} {Reading wikipedia to
  answer open-domain questions}.
\newblock In \emph{ACL}.

\bibitem[{Choi et~al.(2018)Choi, He, Iyyer, Yatskar, tau Yih, Choi, Liang, and
  Zettlemoyer}]{choi2018quac}
Eunsol Choi, He~He, Mohit Iyyer, Mark Yatskar, Wen tau Yih, Yejin Choi, Percy
  Liang, and Luke Zettlemoyer. 2018.
\newblock Quac: Question answering in context.
\newblock In \emph{{EMNLP}}.

\bibitem[{Devlin et~al.(2018)Devlin, Chang, Lee, and
  Toutanova}]{devlin2018bert}
Jacob Devlin, Ming-Wei Chang, Kenton Lee, and Kristina Toutanova. 2018.
\newblock Bert: Pre-training of deep bidirectional transformers for language
  understanding.
\newblock \emph{CoRR}, abs/1810.04805.

\bibitem[{Dunn et~al.(2017)Dunn, Sagun, Higgins, G{\"{u}}ney, Cirik, and
  Cho}]{dunn2017searchqa}
Matthew Dunn, Levent Sagun, Mike Higgins, V.~Ugur G{\"{u}}ney, Volkan Cirik,
  and Kyunghyun Cho. 2017.
\newblock Searchqa: {A} new q{\&}a dataset augmented with context from a search
  engine.
\newblock \emph{CoRR}, abs/1704.05179.

\bibitem[{Fan et~al.(2018)Fan, Grangier, and Auli}]{fan2018controllable}
Angela Fan, David Grangier, and Michael Auli. 2018.
\newblock \href {http://aclweb.org/anthology/W18-2706} {Controllable
  abstractive summarization}.
\newblock In \emph{ACL Workshop on Neural Machine Translation and Generation}.

\bibitem[{Gehring et~al.(2017)Gehring, Auli, Grangier, Yarats, and
  Dauphin}]{gehring2017convolutional}
Jonas Gehring, Michael Auli, David Grangier, Denis Yarats, and Yann~N Dauphin.
  2017.
\newblock {Convolutional Sequence to Sequence Learning}.
\newblock In \emph{Proc. of ICML}.

\bibitem[{Johnson et~al.(2017)Johnson, Douze, and J{\'{e}}gou}]{johnson2017}
Jeff Johnson, Matthijs Douze, and Herv{\'{e}} J{\'{e}}gou. 2017.
\newblock Billion-scale similarity search with gpus.
\newblock \emph{CoRR}, abs/1702.08734.

\bibitem[{Joshi et~al.(2017)Joshi, Choi, Weld, and
  Zettlemoyer}]{joshi2017triviaqa}
Mandar Joshi, Eunsol Choi, Daniel Weld, and Luke Zettlemoyer. 2017.
\newblock \href {https://doi.org/10.18653/v1/P17-1147} {Triviaqa: A large scale
  distantly supervised challenge dataset for reading comprehension}.
\newblock In \emph{ACL}.

\bibitem[{Kocisky et~al.(2018)Kocisky, Schwarz, Blunsom, Dyer, Hermann, Melis,
  and Grefenstette}]{kocisky2018the}
Tomas Kocisky, Jonathan Schwarz, Phil Blunsom, Chris Dyer, Karl~Moritz Hermann,
  Gabor Melis, and Edward Grefenstette. 2018.
\newblock \href {http://aclweb.org/anthology/Q18-1023} {The narrativeqa reading
  comprehension challenge}.
\newblock \emph{TACL}.

\bibitem[{Lin(2004)}]{lin2004rouge}
Chin-Yew Lin. 2004.
\newblock \href
  {https://www.microsoft.com/en-us/research/publication/rouge-a-package-for-automatic-evaluation-of-summaries/}
  {Rouge: a package for automatic evaluation of summaries}.
\newblock In \emph{ACL Workshop on Text Summarization Branches Out}.

\bibitem[{Liu et~al.(2018)Liu, Saleh, Pot, Goodrich, Sepassi, Kaiser, and
  Shazeer}]{liu2018generating}
Peter~J. Liu, Mohammad Saleh, Etienne Pot, Ben Goodrich, Ryan Sepassi, Lukasz
  Kaiser, and Noam Shazeer. 2018.
\newblock \href {https://openreview.net/forum?id=Hyg0vbWC-} {Generating
  wikipedia by summarizing long sequences}.
\newblock In \emph{ICLR}.

\bibitem[{Nguyen et~al.(2016)Nguyen, Rosenberg, Song, Gao, Tiwary, Majumder,
  and Deng}]{nguyen2016ms}
Tri Nguyen, Mir Rosenberg, Xia Song, Jianfeng Gao, Saurabh Tiwary, Rangan
  Majumder, and Li~Deng. 2016.
\newblock \href
  {https://www.microsoft.com/en-us/research/publication/ms-marco-human-generated-machine-reading-comprehension-dataset/}
  {Ms marco: A human generated machine reading comprehension dataset}.
\newblock \emph{CoRR}.

\bibitem[{Ott et~al.(2018)Ott, Edunov, Grangier, and Auli}]{ott2018scaling}
Myle Ott, Sergey Edunov, David Grangier, and Michael Auli. 2018.
\newblock Scaling neural machine translation.
\newblock In \emph{{WMT}}, pages 1--9. Association for Computational
  Linguistics.

\bibitem[{Paulus et~al.(2017)Paulus, Xiong, and Socher}]{paulus17}
Romain Paulus, Caiming Xiong, and Richard Socher. 2017.
\newblock A deep reinforced model for abstractive summarization.
\newblock \emph{arXiv preprint arXiv:1705.04304}.

\bibitem[{Radford et~al.(2018)Radford, Narasimhan, Salimans, and
  Sutskever}]{radford2018improving}
Alec Radford, Karthik Narasimhan, Tim Salimans, and Ilya Sutskever. 2018.
\newblock \href
  {https://s3-us-west-2.amazonaws.com/openai-assets/research-covers/language-unsupervised/language_understanding_paper.pdf}
  {Improving language understanding by generative pre-training.}

\bibitem[{Rajpurkar et~al.(2018)Rajpurkar, Jia, and Liang}]{rajpurkar2018know}
Pranav Rajpurkar, Robin Jia, and Percy Liang. 2018.
\newblock \href {http://aclweb.org/anthology/P18-2124} {Know what you don't
  know: Unanswerable questions for squad}.
\newblock In \emph{ACL}.

\bibitem[{Rajpurkar et~al.(2016)Rajpurkar, Zhang, Lopyrev, and
  Liang}]{rajpurkar2016squad}
Pranav Rajpurkar, Jian Zhang, Konstantin Lopyrev, and Percy Liang. 2016.
\newblock \href {https://doi.org/10.18653/v1/D16-1264} {Squad: 100,000+
  questions for machine comprehension of text}.
\newblock In \emph{EMNLP}.

\bibitem[{Reddy et~al.(2018)Reddy, Chen, and Manning}]{reddy2018coqa}
Siva Reddy, Danqi Chen, and Christopher~D Manning. 2018.
\newblock Coqa: A conversational question answering challenge.
\newblock \emph{arXiv preprint arXiv:1808.07042}.

\bibitem[{Sennrich et~al.(2016)Sennrich, Haddow, and
  Birch}]{sennrich2016neural}
Rico Sennrich, Barry Haddow, and Alexandra Birch. 2016.
\newblock Neural machine translation of rare words with subword units.
\newblock In \emph{ACL}.

\bibitem[{Seo et~al.(2017)Seo, Kembhavi, Farhadi, and
  Hajishirzi}]{seo2017bidirectional}
Minjoon Seo, Aniruddha Kembhavi, Ali Farhadi, and Hannaneh Hajishirzi. 2017.
\newblock Bidirectional attention flow for machine comprehension.
\newblock In \emph{ICLR}.

\bibitem[{Tombros and Sanderson(1998)}]{tombros1998advantages}
Anastasios Tombros and Mark Sanderson. 1998.
\newblock \href {http://doi.acm.org/10.1145/290941.290947} {Advantages of query
  biased summaries in information retrieval}.
\newblock In \emph{SIGIR}.

\bibitem[{Trischler et~al.(2017)Trischler, Wang, Yuan, Harris, Sordoni,
  Bachman, and Suleman}]{trischler2017newsqa}
Adam Trischler, Tong Wang, Xingdi Yuan, Justin Harris, Alessandro Sordoni,
  Philip Bachman, and Kaheer Suleman. 2017.
\newblock \href {http://aclweb.org/anthology/W17-2623} {Newsqa: A machine
  comprehension dataset}.
\newblock In \emph{ACL Workshop on Representation Learning for NLP}.

\bibitem[{Vaswani et~al.(2017)Vaswani, Shazeer, Parmar, Uszkoreit, Jones,
  Gomez, Kaiser, and Polosukhin}]{vaswani2018attention}
Ashish Vaswani, Noam Shazeer, Niki Parmar, Jakob Uszkoreit, Llion Jones,
  Aidan~N Gomez, {\L}ukasz Kaiser, and Illia Polosukhin. 2017.
\newblock \href
  {http://papers.nips.cc/paper/7181-attention-is-all-you-need.pdf} {Attention
  is all you need}.
\newblock \emph{NIPS}.

\bibitem[{Voorhees(2003)}]{voorhees2003overview}
Ellen~M. Voorhees. 2003.
\newblock Overview of the {TREC} 2003 question answering track.
\newblock In \emph{TREC}.

\bibitem[{Weissenborn et~al.(2017)Weissenborn, Wiese, and
  Seiffe}]{weissenborn2017making}
Dirk Weissenborn, Georg Wiese, and Laura Seiffe. 2017.
\newblock Making neural qa as simple as possible but not simpler.
\newblock In \emph{CoNLL}.

\bibitem[{Yang et~al.(2018)Yang, Qi, Zhang, Bengio, Cohen, Salakhutdinov, and
  Manning}]{yang2018hotpotqa}
Zhilin Yang, Peng Qi, Saizheng Zhang, Yoshua Bengio, William~W Cohen, Ruslan
  Salakhutdinov, and Christopher~D Manning. 2018.
\newblock Hotpotqa: A dataset for diverse, explainable multi-hop question
  answering.
\newblock \emph{arXiv preprint arXiv:1809.09600}.

\end{thebibliography}
\bibliographystyle{acl_natbib}

\newpage

\clearpage

\appendix

\section{Details of Multitask Training}
\label{appendix:multitask}
The seq2seq multi-task model was trained on a variety of tasks at training time. Each task is specified by a special token to delineate to the model which task it is. Tasks at training time include the following, in the form of (source, target) pairs. ``+'' represents a concatenation of inputs, separated by a special token. 

\begin{itemize}
  \itemsep0em 
  \item (empty, question)
  \item (empty, document)
  \item (empty, answer)
  \item (empty, question + document)
  \item (empty, question + document + answer)
  \item (question, answer)
  \item (question, document)
  \item (question + document, answer)
  \item (question, document + answer)
  \item masked word prediction: 15\% of source words are replaced by a ``[MASK]'' token and the corresponding tokens must be predicted as the target in the correct order
\end{itemize}

\section{Architectural Details}

\subsection{Extractive BidAF}

The BidAF model is trained using the AllenNLP\footnote{\url{https://allennlp.org/}} implementation, using the standard hyper-parameters (specified in the bidaf.jsonnet file\footnote{\url{https://github.com/allenai/allennlp/blob/master/training_config/bidaf.jsonnet}}). We only change the batch size, since a 16GB GPU can only fit one example per batch, and as a result the Adam learning rate has to be changed to $5e-5$. We provide the code to select the target span and sub-sample the input in our data, as well as to convert it to the SQUAD format required by the AllenNLP system.

\subsection{Abstractive Models}

Models are trained with the Adam optimizer with beta values $(0.9, 0.98)$, initial learning rate $1e-07$ with 4000 warmup steps to learning rate 0.0001. We follow the inverse square root learning rate scheduler described in \cite{vaswani2018attention}. Models are trained with a label smoothing value of 0.1. 

Sequence to sequence models are trained with following architecture from \cite{vaswani2018attention}: 6 encoder layers, 6 decoder layers, FFN dimension 4096, 16 attention heads, embedding dimension 1024. Gradient updating is delayed until 32 updates have been processed. Models are regularized with dropout 0.1 and attention dropout 0.1. 

Language models are trained with same parameters described for seq2seq above, with 6 decoder layers. We did not train with 12 decoder layers, as we found the deeper Transformer model was harder to optimize and we achieved worse results compared to a 6-layer language model. 

For generation, models generate a minimum of 200 words and a maximum of 500 words. 

\section{Comparison of Extractive and Abstractive Methods}

Figure~\ref{fig:poor_align_example} displays an example of a generated answer for an example where the source document is of poor quality but the abstractive answer still has strong ROUGE. In comparison, the extractive answer is heavily affected by the poor document quality and derails in topic.

\section{Test/Valid Similarity with Train}

\begin{figure}[h]
    \centering
    \includegraphics[width=0.6\linewidth]{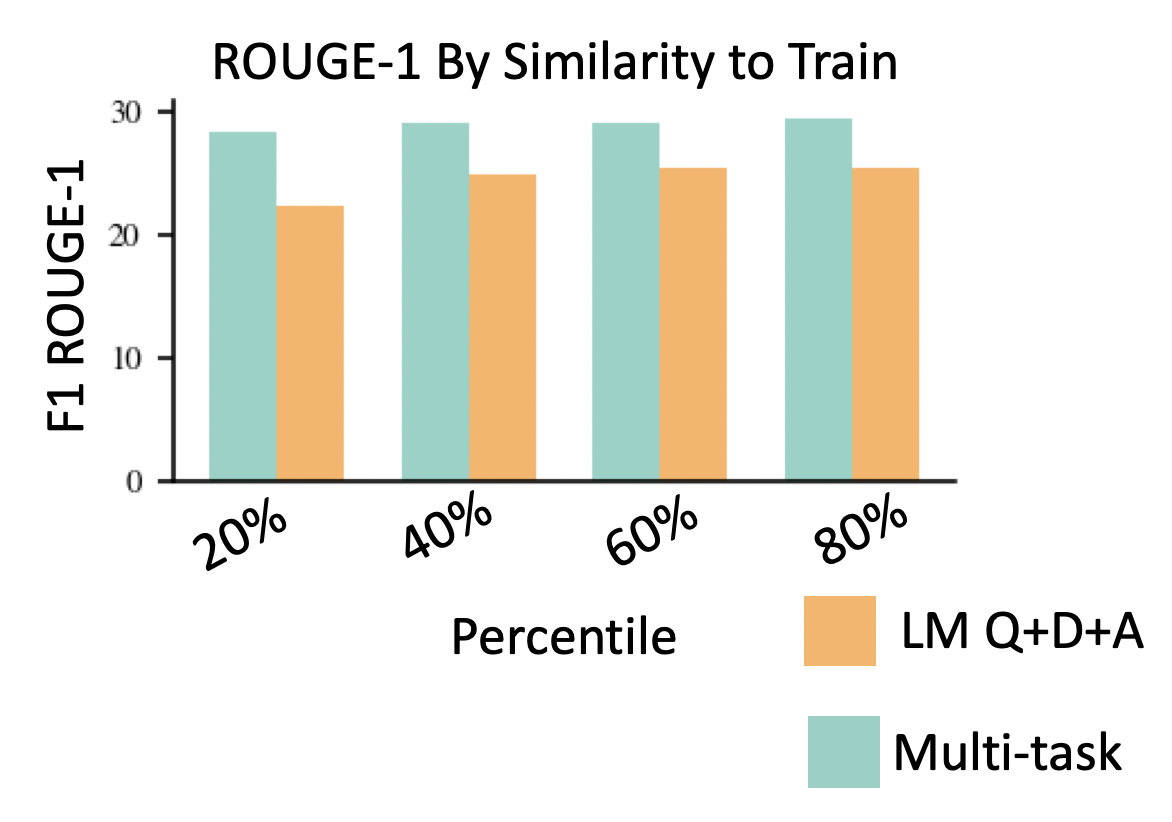}
    \caption{ROUGE of full answer generation is not strongly affected by similarity of the questions in the validation set to questions in the training set.}
    \label{fig:train_similarity_appendix}
\end{figure}

Figure~\ref{fig:train_similarity_appendix} shows the performance of the Multi-task Seq2Seq and LM on Question + Document + Answer by the similarity of the question in the validation set to a question in the training set. The similarity is determined by TFIDF. There is very little effect of answer generation on a question more similar to a training question than less similar. 

\section{Variance in Human Evaluation Studies}
\label{sec:humanvariance}

\begin{figure}
    \centering
    \includegraphics[width=0.6\linewidth]{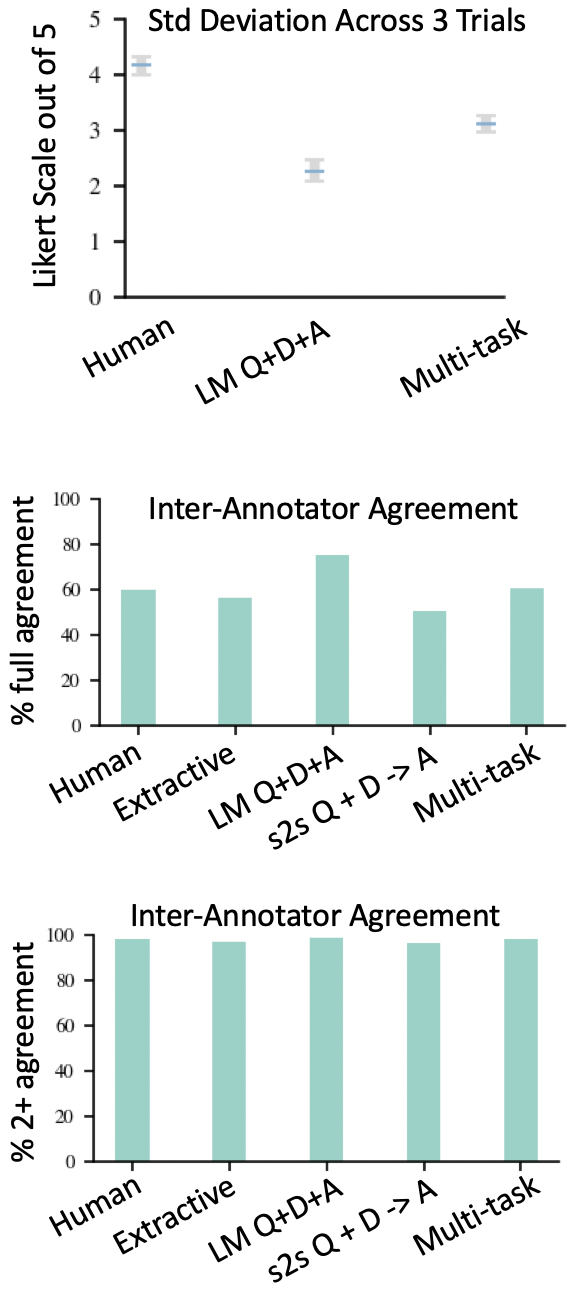}
    \caption{\textbf{Analysis of Human Fluency Study} \textbf{(a)} We analyze the variation between three iterations of the same experiment, conducted on three different days. We evaluate the fluency rating given to the human answers, LM answers, and Multi-task model answers, and find low variation across multiple trials.\\ 
    \textbf{(b)} We calculate the inter-annotator agreement between the three evaluators that score the fluency of each of the models, and find that the \% of time all three annotators agree is highest for the language model. Agreement is calculated using positive (Likert scale scores 4 and 5), neutral (3), and negative (scores 1 and 2). \\
    \textbf{(c)} We calculate the inter-annotator agreement between at least two evaluators for the fluency of each of the models, and find that two annotators usually agree almost 100\% of the time.}
    \label{fig:human_variation_appendix}
\end{figure}

We analyze the variation of our human evaluation study for answer generation fluency in Figure~\ref{fig:human_variation_appendix}. We conduct 3 different trials of the same 100 randomly sampled question-answer pairs from the test set for the selected models. Each trial is conducted on a different day. Our results show that standard deviation across the trials is small and not statistically significant.

Further, each answer is evaluated for fluency by 3 different crowdworkers. Figure~\ref{fig:human_variation_appendix} analyzes the agreement rate between crowdworkers that can choose on a scale of five options. We term ``agreement'' if all workers are positive, negative, or neutral about the answer fluency. We show that all three crowdworkers agree around 60\% of the time for most models and almost 80\% of the time for the language model. As the language model generation is significantly less fluent than the other models, most crowdworkers are in agreement. The agreement of at least two of the annotators is almost 100\% for all of our evaluated systems. 

\section{Examples}
\label{appendix:Examples}

We display randomly sampled examples from the training set in Figure~\ref{fig:example_qa_pairs} and examples of answers that do not answer the question in Figure~\ref{fig:example_non_answered} (an estimated 5\% of the dataset). 

To better understand the output of our models, we display example generations randomly sampled from the test set for the multi-task Seq2Seq model (Figure~\ref{fig:multitask_answers}) and the Extractive BidAF model (Figure~\ref{fig:bidaf_answers}). We additionally show a set of poor generations for the multi-task Seq2Seq model (Figure~\ref{fig:poor_multitask_answers}) that display a representative set of problems for this abstractive model. 

\begin{figure*}[t]
  \small
  \rule{\linewidth}{1pt}

    \noindent \textbf{Why do Oreos soften when they get stale, but most other foods (like bread) harden?}  It has to do with how much moisture is in the food compared to how much moisture is in the air. Oreo? Drier than air, they gain moisture and turn soft. Loaf of bread? More moist than air, it loses moisture and dries up. Kitchen hack: put a slice of bread in with cookies to keep the cookies fresher a little longer. \\

    \noindent \textbf{Why is it that cracks seemingly spread on their own?}  Because of how the mechanics of the crack work. The stress in a material is related to the radius of corners, that is, rounded corners, like the bottom of the letter U, have less stress than sharp corners, like a V. Sharp corners are actually referred to in mechanical engineering as ``stress raisers''. And what's sharper than the edge of a crack? This means that even stress from thermal expansion of the material can be enough to cause the crack to spread. This is especially true of glass, whose brittle composition is quite susceptible to catastrophic failures. \\
    
    \noindent \textbf{Discrete Math: reflexive, symmetric, antisymmetric, transitive}  A *binary relation* is a function of two variables that is either true or false. I'm going to use Rxy to notate the relation R applied to x and y in that order. A relation is *reflexive* if Rxx is true; that is, if both variables are the same the relation always holds. A relation is *symmetric* if Rxy = Ryx; that is, switching the order of the arguments doesn't change the value. A relation is *antisymmetric* if Rxy never equals Ryx for distinct x and y; that is, switching the order of the arguments always changes the value (unless the arguments are the same, in which case it obviously can't). A relation is *transitive* if Rxz is true whenever Rxy and Ryz are. Equality is the simplest example of this; if you have x = y and y = z, then you can conclude that x = z. \\ 
    
    \noindent \textbf{Why does bashing my crt tv make it work?}  There are several reasons why ``percussive maintenance'' can work. The most likely is that there's simply a poor electrical connection somewhere, and banging on it gets it into contact. Once the electricity starts to flow, things heat up and expand a bit, maintaining the connection. Until you turn it off and everything cools off again.\\ 
    
    \noindent \textbf{Is it more efficient to leave the AC running on auto at 74F (in 85 degree whether) or turning it off when leaving the house, and turning it back on when returning when the ambient temp in the apartment is 85?}   Turn it off, you will use less power. Thermodynamics tells us that heat loss is proportional to temperature difference, so if you let your house warm up the heat from outside enters more slowly. Essentially the product of time and temperature difference is your cooling energy. There is a minor subtlety with maintenance and life cycle, your AC unit may not be designed for continuous duty, so long cool down cycles could be hard on it. Not sure if that is the case in your unit, seems like a bad way to design anything but it could be. Edit: one non-thermodynamic factor is humidity and mold, which will be different at a constant temperature vs a cycling temperature. 
  \rule{\linewidth}{1pt}
  \caption{ Examples of Question-Answer Pairs randomly sampled from the training set}
  \label{fig:example_qa_pairs}
\end{figure*}

\begin{figure*}[t]
  \small
  \rule{\linewidth}{1pt}

    \noindent \textbf{The differences between Wii, PS, Xbox, etc., and which is considered the best system. I'm 40, out of the loop, and have to buy one for Christmas.}  If he is 7 go for the Wii its /technically/ more kid friendly and has a more varied option of games for that age range... I think \\ 
    
    \noindent \textbf{What is it when everything just looks really bright/your eyes white out for a moment, then goes back to normal?}   What is it? Time to see a doctor. \\
    
    \noindent \textbf{Neurologically what happens someone becomes insane} That is waaaaaaaay too broad a question to possibly answer since ``insane'' covers a multitude of different conditions and disorders, all with their own causes and that's even assuming that we know the causes in the first place. \\

    \noindent \textbf{If my spaceship could withstand the pressures of Jupiter, what would a trip through the planet be like?}   Read the novel ``2010'' by Arthur C. Clarke. There's a chapter that explains this far better than I could.\\

    \noindent \textbf{If person A saves person B's life through CPR but person B's rib gets broken during chest compressions, why is person B able to sue person A? Why doesn't the judge through the case out of court?}   I know Good Samaritan laws play into this somehow, but I can't remember how. I think that if you have the proper licensing, then they can't sue you. Or maybe it's the other way around. Sorry, this response has become useless \\

  \rule{\linewidth}{1pt}
  \caption{ Examples of Answers in the training set that do not answer the question}
  \label{fig:example_non_answered}
\end{figure*}

\begin{figure*}
  \small
  \rule{\linewidth}{1pt}
    \noindent \textbf{Question:} Why do TV shows hide logos ? \\ 
    \\
    \noindent \textbf{Document:} Why is that? Why do logos get blurred on television, or in film, or even in music videos? The answer, it turns out, is complicated, but is mostly about money in various forms. A whole lot of dysfunction here. Quick Pick: HBO TV Shows Pick the missing word in the title of these HBO shows. September is season premiere month so these logos should be fresh in your mind. TV Shows Venn Diagram II Can you click on the most accurate section of the Venn Diagram for each of the following TV Shows? [...] \\
    \\
    \noindent \textbf{ELI5 Answer:} nothing is free. In most cases, it is a prop for the show, but because apple did NOT pay them for the product placement, the show isn't going to give it away. In other cases, apple may not want their brand used in association with that media.\\
  \rule{\linewidth}{1pt}
    \noindent \\ \textbf{Extractive:} The answer, it turns out, is complicated, but is mostly about money in various forms. There is only one reason they do not trust it and that is fear of loosing their money. All the seals or say reassurance and credibility logos assure your user that a website is safe, protected and personal information is safe. A Lot Of New Disney TV Shows Disney is finally looking to break into the world of TV shows with this streaming service and that's reportedly going to include "episodic spinoffs" of beloved franchises like High School Musical (that's right, it's coming back\ldots), Monsters Inc., and The Muppets.\\
    \\
    \noindent \textbf{Seq2Seq Multi-task:} It's because they're paying for the right to use the show. If you're a TV show, you're paying to have the show shown for free, and you're not paying for a license to use that show for free. the reason they don't have to pay for the license is because the show doesn't want to be associated with the show, and the show isn't paying for it. It's not the show's fault, it's the company that makes the show pay for it, and it's not their fault.\\
    \rule{\linewidth}{1pt}
  \caption{
    Example of document with low document-answer overlap where the abstractive system still achieves a high ROUGE score. The extractive baseline correctly identifies where the question is reformulated in the source document and selects the following sentence, but then goes on to confuse different types of logos and goes off topic entirely with the Disney analysis.
    }
  \label{fig:poor_align_example}
\end{figure*}

\begin{figure*}[t]
  \small
  \rule{\linewidth}{1pt}

    \noindent \textbf{Who controls what the president is allowed to do? But then who controls what they're allowed to do? Etc ..}  The president is the commander-in-chief of the armed forces. He is the head of the executive branch. He controls the military, the military and the military. The president controls the army, the navy, the air force, and the army. The President is the president. He has the authority to do whatever he wants, but he doesn't have the power to do all he wants. He doesn't want to do anything about it. He wants to do something about it, but it doesn't really matter what the president is doing. He's not the president, he's just a figurehead, and he's the president of the United States. It's not really a matter of who controls the government, it's the people who control the government. There are a lot of people who don't like the President, but they don't really like him. There's a lot more to it than that, but that's the basic gist of it. The US government is controlled by the US government, but not by the federal government. The federal government is the one that controls most of the US military, but there are a few that don't. They are the ones that control most of US military operations, and they are not the ones who control most. \\
    
    \noindent \textbf{What makes the stomach ``growl''?} The sound of your stomach growling is caused by the movement of air in your stomach. The sound you hear when your stomach is empty is called peristalsis. This is the sound that your stomach makes when it is empty. When you are hungry your stomach will make the sound of the growling sound, but it will not be as loud as the sound you are making when you are not hungry. It is the same thing that makes your stomach growl when it's empty. You are not actually hungry, but you are actually contracting the muscles in your intestines to make the noise. The growling is the result of the muscles contracting and contracting to make a loud noise. It's the same reason why you feel the urge to eat when you're hungry. Your stomach is trying to empty itself, so it makes the noise as if you aren't hungry. The same thing happens with your stomach when you have a cold. The muscles in the stomach are contracting and making the noise, but they are not contracting. They are contracting, but not contracting, so they aren't contracting. The noise you feel when you feel it is the stomach making is your stomach making the sound. Source : I am a gastroenterologist, and I have a lot of experience with gastroenterology. \\ 

    \noindent \textbf{How does a de-humidifier work?} A humidifier works by heating the air in the room. When the air is warm, it cools down. When it is cold, it warms up. This cools the air around it, which cools it down. This is why you can see a lot of cool things in a room if you look at it from a distance. The air around the room is warmer than it is, so it rises. This causes it to rise, and cool down. The same thing happens with a humidifier. The water in the air condenses out of the air, and condenses back into liquid water. This water condenses back to liquid water, which is what you see when you blow on a hot air balloon. The cold air cools down, and the warm air condenses back out of it, making the air colder. It's the same thing as a humidifier, except that it doesn't have any water in it, it just has a bunch of tiny droplets in it that make up the air. It does have some water, but it has a lot more in it than it does with a cold air balloon, so the air gets colder. The downside to a humidifier is that it isn't as efficient as a regular humidifier, and it is much more efficient than a regular one.\\ 

    \noindent \textbf{Could anyone explain the suffix nomenclature for diseases/conditions/ailments? Ex. -itis vs -osis vs -algia} There are a few different types of suffix. -ous suffix comes from the Latin ``-ary'', which means ``-ate''. The suffix ``-er'' comes from a Latin root that means ``something'' or ``something.''  ``-ous'' is a suffix that is used to refer to things that are related to something that is related to that thing. For example, ``-es'' means ``a''. So ``-e'' is the suffix that comes from ``-ide'', and ``-o'' is ``-ic''. In this case, the ``-i'' is from the Greek ``-os'', so ``-ia'' is just the Latin root of ``-ian''. It's the same way that ``-s'' is also the suffix for ``-ies'', but it 's not the same thing as ``-y''. A lot of the words that are used in the English language don't have the same suffix, so they are used differently. In English, the word ``-us'' has a different suffix than ``-u''. This means that the -ic suffix is more likely to be used in English than the -ian suffix . In other words, it's more likely that the suffix will be used more often than it is in English.\\ 

    \noindent \textbf{How do live visualization websites work?}  I'm not sure what you're referring to, but I'll give it a shot. First of all, you have to understand what they mean by the term `` visualization''. Basically, they use a technique called ``visualization.'' Basically, you take a picture of a room, look at it, and compare it to what it looks like to you. Then, you look at the room, and see what the room looks like, and then compare that to what you see in the room. You can then use the visualization to figure out what it actually looks like. For example, let's say you 're in a room with a camera, and you want to know what the camera looked like. You know that the camera looks like the camera, but you don't know what it looked like, so you use it to look at what it was like to see what you were looking at. Then you use the technique to find out what the person was looking at, and figure out how to use that to get a sense of what they were actually looking at at. The visualization is pretty simple, but it's a lot more complicated than that, so I'll try to explain it in a bit more detail. First, you need to understand how the visualization works. The first thing to understand is that it's very easy to use. Let's say that you're looking at a room and you're really interested in the camera. You want to be able to see how far away the camera is from you, so that you can figure out the distance between the camera and your camera. The second thing to remember is that the visualization isn't very accurate. It's just a way of looking at things that aren't directly related to each other. For instance, if you're interested in learning how to read a book, you might be interested in reading a book. You might also want to read the book in a way that is easy to understand, and it's easy to get used to it.
  \rule{\linewidth}{1pt}
  \caption{ Example Question-Generated Answer pairs from Seq2Seq multi-task model, randomly sampled from the test set
    }
  \label{fig:multitask_answers}
\end{figure*}

\begin{figure*}[t]
  \small
  \rule{\linewidth}{1pt}

    \noindent \textbf{Why doesn't the color red compress well in JPEGs?} PNG will take longer to load on a website, but it’s sometimes simply worth it if it means the difference between a good or bad image. So with all this hate toward JPEGs, you might be asking yourself why the format continues to even exist when so many better options are available. Also important to note, JPEGs do more than compress the file, they also lose color and contrast range. These numbers and ratios are examples for the sake of easy explanation, but lets say a picture has 100 colors and 100 contrast points. Straight out of the camera, JPEGs often look much more vibrant than raw les, because the colors have been enhanced and sharpening applied in-camera. If you need to archive a large number then you could try placing them in a zip file, but you probably won't save more than 5\%.\\ 
    
    \noindent \textbf{When reading weather reports and it says 50\% chance of rain, what does that actually mean and how is it calculated?} I have always maintained this is a confusing concept and it’s the main reason that I will rarely if ever use a percent chance in a forecast. When they say there is a 50\% chance of rain, does that mean that there is a 50\% chance it is not going to rain? Then, why does it always rain when the chance of rain is 50\%? So, maybe the 50\% chance means that it will rain on only 50\% of the land while the other 50\% rains on the water. This is important to keep in mind because when making claims about the impact of global warming, you need to look at the big picture, not just the last 150 years. Well, there are two input variables you have to keep in mind: first, the geographic location --- where you’re looking for a forecast, and second, the time window you’re looking at.\\  

    \noindent \textbf{Why does my skin turn a paler white when pressed? Kinda random. Always wanted to know.} There is a darker shade, but the shade Sunkissed is perfect for the lighter skin wearers. It doesn't irritate eyes, and it's gentler on skin than some of their other powders --- it's also very finely milled and thick enough that you can use it as a foundation and it covers even dark broken capillaries. What I don't like: This is very light peach when it starts out, and it doesn't turn paler on skin; it also oxidizes. It's a light peach when it starts out, and then it turns darker. If you are unsure if you have cool skin tone, check if you have bluish coloured veins inside your wrist (just under your forehand). Spots or a rash that do not fade under pressure will still be seen when the side of a clear drinking glass is pressed firmly against the skin.\\  

    \noindent \textbf{Can psychoactive drugs trigger schizophrenia in people who are at a genetic disposition to the disease already? If so, how high is the risk for different family members that have it? Do you have a higher chance if a parent has it or if a grandparent has it? What about cousins? Aunts and Uncles?} The identical twin of a person with schizophrenia is most at risk, with a 40 to 65 percent chance of developing the disorder. Some doctors think that the brain may not be able to process information correctly; and it is believed that genetic factors appear to play a role , as people who have family members with schizophrenia may be more likely to get the disease themselves. As Schizophrenia has a tendency to run in families, scientists already know there is a genetic link but that doesn’t mean that if you do have someone in your family that has Schizophrenia that you will too, neither does it mean that if you don’t, you won’t, so there are other factors involved too. At the moment people with Schizophrenia are usually prescribed anti-psychotic medication, some of which can carry unpleasant side effects. If you have a pre-existing risk for schizophrenia (which most people at risk are unaware of), there’s a much higher chance that using cannabis will trigger a schizophrenic episode. Again, it is extremely important to note that this risk applies primarily to people who are already at risk of developing schizophrenia.\\ 

    \noindent \textbf{Why has the Canadian dollar gone down in value over the few years?} So far in 2016, the Canadian dollar has lost a lot of value. The days of the Canadian dollar at parity with the US dollar are long gone. A lot of that increase in book value is because of the loss of value of the Canadian dollar. What we have to remember however is that it is not really the Canadian dollar that has gone up in value, it is the American dollar which has gone down. Since the beginning of the Iraq War the American economy has stumbled and one could make the argument that it is because there is no one at the wheel. This means Canadians can now come into the US and get a great deal on just about anything , and they are buying up real estate, to the tune of more than \$20 billion a year. \\
    
    \noindent \textbf{How do they make Jello with pineapple bits in it if pineapple ruins Jello? I have a store-bought cup of generic Jello with pineapple bits in it. According to all my google searching and middle school science classes, this shouldn't be possible because pineapple enzyme just destroys Jello's jellystuff. So what is this thing in my cupboard?} Enzymes: what’s in your spit fruit and jello an hour before the lab begins it is possible to make jello with canned pineapple chunks but not fresh. Well, just as a little science experiment, I stuck a piece of fresh, uncooked pineapple in a cup of the jello, and this is what it looked like after about 20 minutes. In the cooking world, on the other hand, bromelain is regarded as the enemy of the gelatin dessert if you use fresh pineapple in gelatin, the enzyme eats the. Experiment demonstrates denaturation of an enzyme using an enzyme called catalase which is found in the pineapple and jelly experiment background. In this lab students will investigate a few of the different enzymes from our body pineapple contains the enzyme bromelain, and should break down the jello. Well, to begin with, it's not true you can't put any pineapples in jello jello is alive --- at least according to a 1974 experiment performed by dr.
    
  \rule{\linewidth}{1pt}
  \caption{ Example Question-Generated Answer pairs from the extractive BidAF model, randomly sampled from the test set
    }
  \label{fig:bidaf_answers}
\end{figure*}

\begin{figure*}[t]
  \small
  \rule{\linewidth}{1pt}

    \noindent \textbf{Why is ``PC Gamer'' Hardware so Flashy?}   It's because they don't want to. They want to make sure that they have a decent amount of battery life, and that they can make a decent battery life. They also want to be able to make a small battery that will last for a very long time, and make sure they have enough power to last a long time. That said, there are a lot of people who don't know what they're talking about, and don't really know what's going on. There's also the fact that they're not going to make it to the end of the day, and it's going to be a lot harder for them to make the battery last longer than it is for the average person. It's not like they 're going to have to make their own battery last a lot longer than they are for a PC, but they're still going to get a lot more battery life than they're for a gaming PC, so it's a lot easier to make them last longer, and they're also going to need a lot less power than they were for a desktop PC, which means they'll need more power to run the game, and so they'll have to have more power for the game to last longer. That's why it's so important to keep in mind that there 's a ton of people out there who aren't very good at it, and there's no reason to be afraid of it. \\

    \noindent \textbf{ What are the benefits of an American city being it's own county? i.e. San Francisco}  There are a lot of benefits to an American city being ``own'' a county. It's not like there's a whole lot of money to be made in a city, but it's a lot more complicated than that. There's also the fact that there are **lots*** of people who aren't part of the city, and there aren't **many** people who ** are part of that city. There are also **a lot** more people who don't want to be part of a city than there are people who just want to live there. There is also a **lot* of **money** in that city, so it's easier for them to get there than it is for the people who want to stay there. This is why there are so many people in San Francisco, and why it's so hard to find a city that doesn't have a whole bunch of people in it. **Edit:** [Here's a link] (URL\_0) to an article that explains it better than I can. I'm not sure if this is the right place to ask this, but I'm sure someone else can give you a better answer than me, so I 'll give it a shot. \\
    
    \noindent \textbf{How do cruise ships get wifi in the middle of the ocean?}  There are a couple of ways to do this. One is to turn on the radio and turn off the wifi. The other is to make sure that the wifi is turned on. If you turn the wifi on, the radio will turn on and the wifi will turn off. This is done by turning on the wifi and turning off the radio. If the WiFi is turned off, then the WiFi will turn back on, but the radio won't turn on. There are also other ways to turn off wifi, such as turning on wifi, or turning off wifi. I'm not sure how to explain this to a 5 year old, but I'll try to explain it to a five year old. I hope this helps! Edit: I forgot to mention that WiFi is not a wireless network, it is a network of wireless networks. WiFi networks are connected to the internet, but they are not connected to WiFi networks. The internet is connected to a WiFi network, but it isn't connected to Wifi networks. They are connected by WiFi networks, but WiFi networks can not be connected to wifi networks. So, if you turn on WiFi, the WiFi network will be turned on and off. The WiFi network can be turned off by turning off WiFi, but not by turning it back on. \\
    
    \noindent \textbf{How do they know exactly how tall Mt. Everest is?}  There are a lot of different ways to measure the height of an object. For example, you can measure the distance from the top of a mountain to the bottom of the mountain, and you can calculate the distance between the top and the mountain. The easiest way to do this is by measuring the angle between the mountain and the ground, and then measuring the height between the ground and the top. You can also measure the angle of the ground from the bottom to the top, and the height from the ground to the ground. The most common method is to use a ruler, which is a ruler that measures the length of the base of a piece of paper. The height of the top is measured in meters, and it's measured in feet. The simplest way to measure height is by using a ruler. A ruler is a measure of the distance to the base, and a ruler measures the height at which the base is measured. A good ruler is also a good ruler, but it's not a good one. It's a good measure of how far away a mountain is, and how far it is away from the base. The same is true for measuring height, but you can't measure it with a ruler because it's too far away. \\
    
    \noindent \textbf{How/where was the first language invented?}  It wasn't until the late 19th century that the first language was invented. It was just a way for people to communicate with each other and to communicate. For example, if you were to talk to someone, they would talk to yo , but you wouldn't know what they were talking about. If you were talking to them, they'd talk to each other, but they'd never know what you're talking about, so they wouldn't be able to tell you what they're talking to. The same thing happens with languages. For instance, you might say ``I 'm talking to you'', but if you're not talking to me, you won't know how to say it. You might say, ``I want to know you,'' but you don't know where you're going to be talking to, so you might not know what to say. But you might also say ``you want to understand me,'' and so on. And so on and so forth. Eventually, people started to learn how to communicate, and eventually, they started to figure out how to use their own language to communicate in a way that they couldn't understand, and then they developed a new language that they didn't understand. It's like how you can tell a friend that you're in the middle of a conversation, and you can't tell them how to tell them that they're in a conversation. 
  \rule{\linewidth}{1pt}
  \caption{ Example poor quality Question-Generated Answer pairs from the Seq2Seq multi-task model}
  \label{fig:poor_multitask_answers}
\end{figure*}

\end{document}